\documentclass[conference]{IEEEtran}

\IEEEoverridecommandlockouts

\usepackage{graphicx}
\usepackage{caption}
\graphicspath{
    {figures/}
}

\usepackage{amsmath,amssymb,enumerate}

\usepackage{amsthm}
\usepackage{setspace}
\usepackage{booktabs}
\usepackage[usenames,dvipsnames,svgnames,table]{xcolor}
\usepackage{mathtools}
\usepackage{algorithm, algorithmicx, algpseudocode}

\usepackage{blindtext}
\usepackage{gensymb}
\usepackage{xparse}
\usepackage{lipsum}
\usepackage{mathrsfs}
\usepackage[mathscr]{euscript}
\usepackage{times}
\usepackage{cite} 
\usepackage{multicol}
\usepackage{multirow}
\definecolor{darkgreen}{rgb}{0,0.6,0}
\usepackage[bookmarks=true,colorlinks=true,pdfpagemode=UseNone,citecolor=darkgreen,linkcolor=black,urlcolor=BrickRed]{hyperref}
\usepackage[caption=false,font=footnotesize]{subfig}
\usepackage{amsfonts}
\usepackage[utf8]{inputenc}
\usepackage[T1]{fontenc}
\usepackage{textcomp}
\usepackage{arydshln}
\usepackage{balance}

\renewcommand{\thefigure}{\arabic{figure}}

\definecolor{note}{rgb}{0.1,0.1,1}
\definecolor{rephase}{rgb}{0.15,0.7,0.15}
\definecolor{bag}{rgb}{0.6,0.6,0.2}

\makeatletter
\renewcommand*\env@matrix[1][c]{\hskip -\arraycolsep
  \let\@ifnextchar\new@ifnextchar
  \array{*\c@MaxMatrixCols #1}}
\makeatother

\newcommand{\transpose}{\mathsf{T}}

\makeatletter
\newcommand{\mathleft}{\@fleqntrue\@mathmargin0pt}
\newcommand{\mathcenter}{\@fleqnfalse}
\makeatother

\def\BibTeX{{\rm B\kern-.05em{\sc i\kern-.025em b}\kern-.08em
    T\kern-.1667em\lower.7ex\hbox{E}\kern-.125emX}}
\begin{document}

\title{Legged Robot State Estimation within Non-inertial~Environments\\
}

\author{Zijian He$^{1}$, Sangli Teng$^{2}$, Tzu-Yuan Lin$^{2}$, Maani Ghaffari$^{2}$, Yan Gu$^{1}$
\thanks{$^{1}$ Purdue University, West Lafayette, IN 47907, USA.
	E-mails: \{\tt \small he348,yangu\}@purdue.edu.}
\thanks{$^{2}$ University of Michigan, Ann Arbor, MI 48109, USA.
    E-mail: \{\tt \small sanglit,tzuyuan,maanigj\}@umich.edu.}}

\pagestyle{plain}

\maketitle

\begin{abstract}
This paper investigates the robot state estimation problem within a non-inertial environment.
The proposed state estimation approach relaxes the common assumption of static ground in the system modeling. 
The process and measurement models explicitly treat the movement of the non-inertial environments without requiring knowledge of its motion in the inertial frame or relying on GPS or sensing environmental landmarks.
Further, the proposed state estimator is formulated as an invariant extended Kalman filter (InEKF) with the deterministic part of its process model obeying the group-affine property, leading to log-linear error dynamics.
The observability analysis of the filter confirms that the robot's pose (i.e., position and orientation) and velocity relative to the non-inertial environment are observable.
Hardware experiments on a humanoid robot moving on a rotating and translating treadmill demonstrate the high convergence rate and accuracy of the proposed InEKF even under significant treadmill pitch sway, as well as large estimation errors.
\end{abstract}

\begin{IEEEkeywords}
state estimation, non-inertial environments, invariant filtering, legged robots.
\end{IEEEkeywords}

\section{Introduction}

Legged robots capable of operating inside a non-inertial environment can benefit a wide range of critical applications such as emergency response, inspection, maintenance, and surveillance while operating on moving public transit vehicles, ships, and airplanes \cite{9108552,iqbal2022drs,gao2022time}.
Enabling this new robot functionality demands reliable robot state estimation within a non-inertial environment.
However, real-world non-inertial environments such as ships and underwater vehicles exhibit continuous and time-varying ground movement with respect to the inertial frame \cite{gao2022invariant}, breaking the common assumption of static ground in existing state estimation methods~\cite{hartley2020contact}.
Further, they are typically GPS-denied and may also be windowless, preventing exteroceptive robot sensors, such as cameras and LiDARs, from accessing environmental landmarks.
This paper focuses on real-time state estimation for legged robot navigation within a GPS-denied and enclosed non-inertial environment.

Various filtering approaches have been created for legged or general ground robot operation in an inertial environment~\cite{hartley2020contact,lin2023proprioceptive}. 
These approaches assume that the ground is static in the inertial frame.
Under this assumption, the absolute velocity of the robot-ground contact point is zero in the inertial frame when there is no relative motion between the robot's leg or wheel and the ground.
Such a pseudo measurement has been used to correct the state estimates through the fusion with inertial~\cite{bloesch2013state}, visual~\cite{6858831}, and leg~\cite{hartley2020contact} or wheel~\cite{wenzel2006dual} odometry based on extended Kalman filters~\cite{maybeck1982stochastic} and its variations~\cite{9840886,sola2017quaternion,barrau2016invariant, yu2023fully}.
Yet, in a non-inertial environment, the static-ground assumption and pseudo measurements are no longer valid.


\begin{figure}[t]
\centering\includegraphics[width=0.99\linewidth]{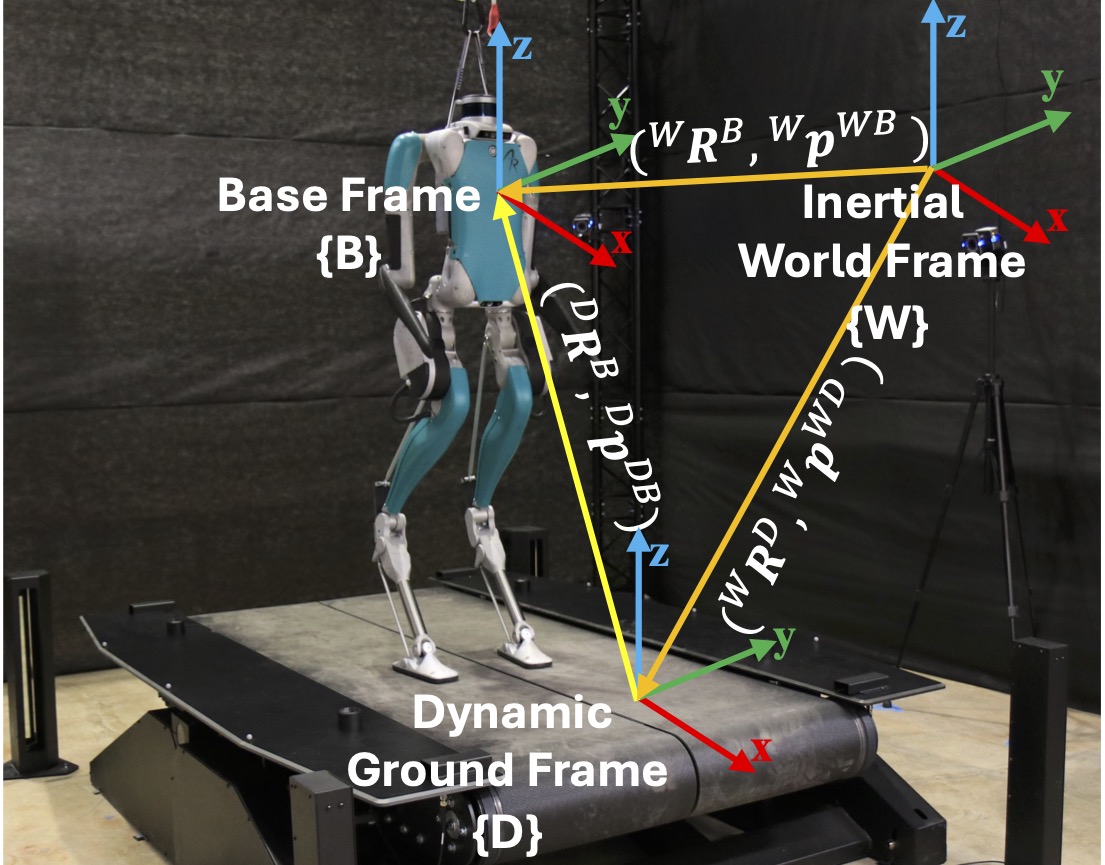}
\caption{Illustration of the reference frames used in the filter derivation.}
\label{reference frame}
\vspace{-0.2 in}
\end{figure}

To relax the zero contact velocity condition, foot or wheel slippage has been treated through visual-inertial odometry with direct measurement of the body velocity in the world frame~\cite{teng2021legged}, and inertial-wheel odometry by rejecting slippage as outliers~\cite{6697236}.
Still, when the ground motion relative to the inertial frame is persistent and multi-directional with a large amplitude, the ignored ground motion acts as significant uncertainty, causing performance degradation~\cite{iqbal2024HTLIP,iqbal2023asymptotic}.

To explicitly handle the ground motion of a non-inertial environment, the static ground assumption has been relaxed \cite{gao2022invariant,gao2021invariant} without relying on GPS or sensing landmarks attached to an inertial environment.
The filter fuses inertial and leg odometry based on the assumption that the absolute ground pose and velocity in the inertial frame are accurately known. 
Nevertheless, such an assumption does not hold in many practical scenarios where the ground movement in the inertial frame cannot be directly sensed or estimated. 

Relative pose estimation of multi-agent systems \cite{7442874}, including those comprising unmanned aerial vehicles, are related to the problem of robot state estimation within a non-inertial environment.
The similarity of these two classes of problems lies in the fact that 
the reference frames attached to moving agents are non-inertial frames.
The pose of those non-inertial frames can be recovered based on data from cameras \cite{6719359}
or
laser scanners~\cite{7442874}.
However, the sampling frequency of these sensors is usually slow or highly costly compared to proprioceptive robot sensors, e.g., inertial measurement units (IMUs) and encoders, and can lead to an overly low estimation rate unsuitable for real-time control and planning.

Beyond state estimation in non-inertial environments, invariant extended Kalman filtering (InEKF)~\cite{5400372, phogat2020invariant, 6160733,barrau2014intrinsic,barrau2019extended} has been introduced to enable fast error convergence under significant errors by exploiting the symmetry reduction for systems evolving on matrix Lie groups.
By the InEKF theory \cite{barrau2016invariant}, if the deterministic unbiased process model satisfies the group-affine property, then there exists an exactly log-linear error dynamics in the Lie algebra. Also, given an invariant observation, the filter is provably convergent under arbitrary initial error, and nonlinear error can be recovered exactly at any time.

InEKF has been applied to solve the state estimation problem for legged \cite{hartley2020contact} and wheeled~\cite{brossard2020ai, lin2023proprioceptive} locomotion on the stationary ground. 
However, the applicability of the InEKF for state estimation in non-inertial environments remains under-explored due to the general reasons mentioned earlier. 

\vspace{-0.05 in}
\subsection{Contributions}

This paper presents an InEKF approach that estimates the relative pose and velocity of a legged robot moving inside a non-inertial environment in the presence of significant estimation errors. 
The proposed filter relaxes the typical assumption of robot state estimation that the ground is static in the inertial frame, and its underlying system models explicitly consider the ground movement without requiring knowledge of the ground motion in the inertial frame.
The key contributions include:
\begin{enumerate}
    \item The standard leg odometry-based measurement model is analytically extended from inertial to non-inertial environments based on the kinetic characteristics of robot movement on an accelerating ground. 
    \item The deterministic part of the process model of the proposed filter is formulated to be group affine, and thus the associated logarithmic error equations are independent of state trajectories and exactly linear. 
    \item Fusing leg and inertial odometry and data returned by an IMU attached to the dynamic ground, the proposed filter renders the robot's relative position, orientation, and velocity observable without requiring input from exteroceptive robot sensors such as cameras and LiDARs.
    \item Hardware experiments on a humanoid robot moving on a pitch and sway treadmill validate the theoretical results.
\end{enumerate}

\subsection{Mathematical Preliminaries and Notations}
Consider a matrix Lie group $\mathcal{G} \subset \mathbb{R}^{n \times n}$.
Its Lie algebra $\mathfrak{g}$ is the tangent space at the group identity element $\mathbf{I}_d$.
The isomorphism, $\left( \cdot \right)^{\wedge}:\mathbb{R}^{\mathrm{dim\mathfrak{g}}} \rightarrow \mathfrak{g}$, maps any vector $\boldsymbol{\xi} \in \mathbb{R}^{\mathrm{dim\mathfrak{g}}}$ to the Lie algebra.
The exponential map of the Lie group, $\exp: \mathbb{R}^{\mathrm{dim\mathfrak{g}}} \rightarrow \mathcal{G}$, is given by
$ \label{matrix exp}
    \exp\left( \boldsymbol{\xi}\right) = \exp_{m}\left( \boldsymbol{\xi}^{\wedge}\right), 
$
where $\exp_{m}(\cdot)$ is the usual matrix exponential. 
For any $\boldsymbol{\xi} \in \mathbb{R}^{\mathrm{dim\mathfrak{g}}}$ and $\mathbf{X} \in \mathcal{G}$, the adjoint matrix $\mathrm{Ad}_{\mathbf{X}}:\mathfrak{g} \to \mathfrak{g}$ performs a change of basis for velocities to account for the change of observing frame, and is defined as:
$
    \left(\mathrm{Ad}_{\mathbf{X}} \boldsymbol{\xi}\right)^{\wedge} = \mathbf{X}\boldsymbol{\xi}^{\wedge} \mathbf{X}^{-1}.
$

We use $\Bar{\left( \cdot \right)}$ to represent the estimated value of the variable $\left( \cdot \right)$,
while $\Tilde{\left( \cdot \right)}$ denotes a measurement of $\left( \cdot \right)$.
The right subscript $t$ of $\left( \cdot \right)_t$ indicates the time $t$.

We use $\{ D \}$, $\{ W \}$, and $\{ B \}$ to denote the reference frames attached to the dynamic ground of the non-inertial environment, the inertial world frame, and the base link of the robot (see Fig.~\ref{reference frame}).
Also, $\mathbf{R}$, $\mathbf{p}$, and $\mathbf{v}$ respectively denote the orientation, position, and velocity of a given object.

The left superscript of a position, orientation, or velocity variable denotes the coordinate system where the variable is expressed.
If the right superscript contains two letters, then the first and the second letters respectively represent the reference frame and the object of interest.
For instance, we use ${}^{D}\mathbf{p}^{DB}$ to represent the relative position of robot's base frame $\{ B \}$ relative to the origin of the dynamic ground frame $\{ D \}$, expressed in $\{ D \}$.
If the right superscript only has one letter, then it represents the object of interest.
For example, ${}^{D}\mathbf{R}^{B}$ denotes the orientation of the robot's base frame $\{ B \}$ with respect to the dynamic ground frame $\{ D \}$.

\section{Problem Formulation}

When a robot navigates inside a non-inertial environment, its planners and controllers need to be informed of the robot's movement state with respect to the non-inertial environment instead of the usual inertial frame. 
However, typical proprioceptive sensors of current robotic platforms (e.g., IMUs and encoders) do not directly measure such states.
Therefore, the proposed filter aims to estimate the orientation, velocity, and position of the robot's base frame within the dynamic ground frame as follows.

\subsection{Sensor Measurements}
The sensors include (a) a robot's IMU mounted at the robot torso (i.e., the base link), which measures the angular velocity and linear acceleration of the base with respect to the base frame $\{ B \}$, and (b) joint encoders, which measure the joint angles $\mathbf{q}_{t}$ of the robot.
Additionally, we consider an external IMU attached to the non-inertial dynamic ground frame $\{ D \}$ whose data is shared with the robot.
This IMU can be placed at any location that is fixed to the dynamic ground.
Such an external sensor setting is general and common as non-inertial platforms such as ships and airplanes are typically equipped with onboard IMUs.

Without loss of generality, we assume that the IMU frames of the robot and the dynamic ground are respectively aligned with the robot's base frame $\{ B \}$ and the ground frame $\{ D \}$.
The joint angle data $\Tilde{\mathbf{q}}_{t}$ returned by encoders is assumed to be corrupted by additive white Gaussian noise. 
The angular velocity and linear acceleration data from the two IMUs at time $t$ are respectively denoted as ${}^i\Tilde{\boldsymbol{\omega}}^{Wi}$ and ${}^i\Tilde{\mathbf{a}}^{Wi}$ with $i \in \{ B,D\}$. 
We assume the sensor data is corrupted by additive white Gaussian noise, ${}^i\mathbf{w}_{t}^{g}$ and ${}^i\mathbf{w}_{t}^{a}$.
For brevity, let
\begin{equation}
\begin{aligned}
        {}^{i}\Tilde{\boldsymbol{{\omega}}}_{t}:={}^{i}\Tilde{\boldsymbol{{\omega}}}_{t}^{Wi}~\text{and}~
    {}^{i}\Tilde{\mathbf{{a}}}_{t}:={}^{i}\Tilde{\mathbf{{a}}}^{Wi}_{t},
\end{aligned}
\end{equation}
and then we can express the sensor data as:
\begin{equation*} \label{system model}
    \begin{split}
        &{}^{i}\Tilde{\boldsymbol{{\omega}}}_{t} = {}^{i}\boldsymbol{{\omega}}^{Wi}_{t} + {}^{i}\mathbf{w}_{t}^{\omega}~\text{and}~
        {}^{i}\Tilde{\mathbf{{a}}} = {}^{i}\mathbf{{a}}^{Wi}_{t} +  {}^{i}\mathbf{w}_{t}^{a},
    \end{split}
\end{equation*}
where ${}^i {\boldsymbol{\omega}}^{Wi}$ and ${}^i {\mathbf{a}}^{Wi}$ are the true angular velocity and linear acceleration being measured. 


\subsection{IMU Motion Dynamics}


We use ${}^{W}\mathbf{R}_{t}^{i}$, ${}^{W}\mathbf{v}_{t}^{Wi}$, and ${}^{W}\mathbf{p}_{t}^{Wi}$ to respectively denote the absolute orientation, velocity, and position of the reference frame $\{ i \}$ with respect to the world frame $\{ W \}$, with $i \in \{ B, D\}$ (see Fig.~\ref{reference frame}).
Then, the IMU dynamics for the frame $\{ i \}$ are~\cite{hartley2020contact}:
\begin{equation} 
\label{process model}
    \begin{split}
    &\frac{d}{dt}\left({}^{W}\mathbf{R}_{t}^{i} \right)= {}^{W}\mathbf{R}_{t}^{i} \left[{}^{i}\Tilde{\boldsymbol{{\omega}}}_{t} - {}^{i}\mathbf{w}_{t}^{\omega}\right]_{\times},
    \\
    &
    \begin{split}
        \frac{d}{dt} \left({}^{W}\mathbf{v}_{t}^{Wi} \right) 
        &=  {}^{W}\mathbf{R}_{t}^{i} \left({}^{i}\Tilde{\boldsymbol{{\mathbf{a}}}}_{t}  - {}^{i}\mathbf{w}_{t}^{a}\right) + \mathbf{g},
    \end{split}
    \\
    &\frac{d}{dt}\left( {}^{W}\mathbf{p}_{t}^{Wi}\right) = {}^{W}\mathbf{v}_{t}^{Wi},
    \end{split}
\end{equation}
where $[ ~ ]_{\times}$ denotes the skew-symmetric matrix of a vector and $\mathbf{g}$ is the gravitational acceleration.

\subsection{Leg Odometry}

We denote the position of the robot's stance foot relative to the robot's base, expressed in the base frame as 
${}^{B}\mathbf{p}^{BF}_{t}$.
Since the robot's joint angles $\mathbf{q}_t$ are directly measurable, we introduce the forward kinematics function ${s}(\mathbf{q}_{t})$ satisfying
    ${}^{B}\mathbf{p}^{BF}_{t}={s}(\mathbf{q}_{t})$, where
${s}(\mathbf{q}_{t})$ is known for a given robot.

The usual measurement model built upon the leg odometry typically assumes that the ground is stationary in the inertial frame, which breaks down for non-inertial environments.
Therefore, we will introduce a new measurement model based on the leg odometry.



\section{Process and Measurement Models}

This section presents the proposed process and measurement models that serve as the basis of the proposed InEKF. 

\subsection{Process Model}

The process model describes the propagation step, i.e., during the time period between successive instants of measurement updates.
For brevity, let $\mathbf{R}_{t}:= {}^{D}\mathbf{R}^{B}_{t}$ and $\mathbf{p}_{t} := {}^{D}\mathbf{p}^{DB}_{t}$.

Given the IMU motion dynamics in Sec. II and the relationship ${}^D \mathbf{R}_t^B = ( {}^W \mathbf{R}_t^D )^\transpose ( {}^W \mathbf{R}_t^B ) $, we obtain the following dynamics of the robot's relative orientation ${}^D \mathbf{R}_t^B$ during the propagation step:
\begin{equation} \label{Rt}
    \frac{d}{dt} \mathbf{R}_{t} = \mathbf{R}_{t} \left[{}^{B}\Tilde{\boldsymbol{{\omega}}}_{t} - {}^{B}\mathbf{w}_{t}^{\omega}\right]_{\times} - \left[{}^{D}\Tilde{\boldsymbol{{\omega}}}_{t} - {}^{D}\mathbf{w}_{t}^{\omega}\right]_{\times} \mathbf{R}_{t}.
\end{equation}

Since the dynamic ground frame $\{ D \}$ translates and rotates in the inertial frame, the dynamics of the robot's relative position during the propagation step are given by:
\begin{equation} \label{pt}
    \frac{d}{dt} \mathbf{p}_{t}  = -\left[{}^{D}\Tilde{\boldsymbol{{\omega}}}_{t}  - \mathbf{w}_{D}^{\omega}\right]_{\times} 
    \mathbf{p}_{t} + \mathbf{v}_{t},
\end{equation}
where $\mathbf{v}_{t}$ is defined as: 
\begin{equation}
    \mathbf{v}_{t}:= ({}^{W}\mathbf{R}_t^{D} )^\transpose \left({}^{W}\mathbf{v}_t^{WB} -  {}^{W}\mathbf{v}_t^{WD}\right).
    \label{v_t}
\end{equation}

Taking the first time derivative of both sides in \eqref{v_t} yields the dynamics model of the state variable $\mathbf{v}_{t}$ as:
\begin{equation} \label{vt}
\begin{aligned}
     \frac{d}{dt} \mathbf{v}_{t} 
        &= -\left[{}^{D}\Tilde{\boldsymbol{{\omega}}}_{t} - {}^{D}\mathbf{w}_{t}^{\omega} \right]_{\times} 
        \mathbf{v}_{t} + \mathbf{R}_{t} \left({}^{B}\Tilde{\boldsymbol{{\mathbf{a}}}}_{t}  - {}^{B}\mathbf{w}_{t}^{a}\right)
        \\
        &\quad - \left({}^{D}\Tilde{\mathbf{a}}_{t}  - {}^{D}\mathbf{w}_{t}^{a}\right).
\end{aligned}
\end{equation}

The state variables $\mathbf{R}_{t}$, $\mathbf{v}_{t}$, and $\mathbf{p}_{t}$ can be expressed on the matrix Lie group $\mathcal{G} \subset \mathbb{R}^{9 \times 9}$ as:
\begin{equation} \label{state_matrix_IMU}
    \mathbf{X}_{t} = 
\begin{bmatrix}
    \mathbf{R}_{t} &\mathbf{v}_{t} &\mathbf{p}_{t}
    \\
    \mathbf{0}_{1,3} & 1 & 0
    \\
    \mathbf{0}_{1,3} & 0 & 1
\end{bmatrix},
\end{equation}
where $\mathbf{0}_{m,n}$ is an $m \times n$ zero matrix.
Here the Lie group $\mathcal{G}$ is the direct isometries group $\mathrm{SE}_{2}(3)$~\cite{barrau2015non}.

Defining the input $\mathbf{u}_t$ to the process model as:
\begin{equation}
            \mathbf{u}_{t} = 
        \begin{bmatrix}
            ( {}^{B}\Tilde{\boldsymbol{{\omega}}}_{t} )^\transpose
            &
            ( {}^{D}\Tilde{\boldsymbol{{\omega}}}_{t} )^\transpose
            &
            ( {}^{B}\Tilde{\mathbf{{a}}}_{t} )^\transpose
            &
            ( {}^{D}\Tilde{\mathbf{{a}}}_{t} )^\transpose
        \end{bmatrix},
\end{equation}
the process models in \eqref{Rt}, \eqref{pt}, \eqref{vt} can be expressed as:
\begin{equation} \label{matrix process model} 
\small
    \begin{split}
        \frac{d}{dt} \mathbf{X}_{t}
         &= -{}^{D}\Tilde{\mathbf{U}}_{t}\mathbf{X}_{t} + \mathbf{X}_{t} {}^{B}\Tilde{\mathbf{U}}_{t} + ({}^{D}\mathbf{w}_{t})^{\wedge} \mathbf{X}_{t} -\mathbf{X}_{t} ({}^{B}\mathbf{w}_{t})^{\wedge}
        \\
        & =: f_{u_{t}}(\mathbf{X}_{t}) + ({}^{D}\mathbf{w}_{t})^{\wedge}\mathbf{X}_{t} - \mathbf{X}_{t} ({}^{B}\mathbf{w}_{t})^{\wedge},
        \end{split}
\end{equation}
where
$
{}^i \mathbf{w}_{t} := 
        \begin{bmatrix}
            ({}^{i}\mathbf{w}_{t}^{g})^\transpose
            &
            ({}^{i}\mathbf{w}_{t}^{a})^\transpose
            &
            \mathbf{0}_{1,3}
        \end{bmatrix}^\transpose
$ and
\begin{equation} 
\label{tildeU}
    \begin{split}
        &{}^{i}\Tilde{\mathbf{U}}_{t} :=
        \begin{bmatrix}
        \left[{}^{i}\Tilde{\boldsymbol{\omega}}_{t} \right]_{\times}& {}^{i}\Tilde{\mathbf{a}}_{t} &\mathbf{0}_{3,1}
            \\
            \mathbf{0}_{1,3} & 0 & 1 
            \\
            \mathbf{0}_{1,3} & 0 & 0
        \end{bmatrix}
    \end{split}
\end{equation}
with $i \in \{ D, B\}$.
\hfill

\vspace{+0.05 in}
\noindent \textbf{Proposition 1:} 
{\it The deterministic part of the system dynamics in \eqref{matrix process model}, i.e., $\frac{d}{dt} \mathbf{X}_t =  f_{u_{t}}\left( \mathbf{X}_t \right)$, is group affine.}

\vspace{+0.05 in}
\noindent {\it Proof}:
From the process model in \eqref{matrix process model}, we know
$f_{u_{t}} \left( \mathbf{X}_t \right)
:=
-{}^{D}\Tilde{\mathbf{U}}_{t}\mathbf{X}_{t} + \mathbf{X}_{t} {}^{B}\Tilde{\mathbf{U}}_{t}$.
Thus, for any $ \boldsymbol{\mathbf{X}}_1, \boldsymbol{\mathbf{X}}_2 \in \mathcal{G}$, we have:
\begin{equation}
    f_{u_{t}} \left( \boldsymbol{\mathbf{X}}_1  \boldsymbol{\mathbf{X}}_2 \right)
    = 
    -{}^{D}\Tilde{\mathbf{U}}_{t}  \boldsymbol{\mathbf{X}}_1 \boldsymbol{\mathbf{X}}_2  + \boldsymbol{\mathbf{X}}_1 \boldsymbol{\mathbf{X}}_2 {}^{B}\Tilde{\mathbf{U}}_{t}.
\end{equation}

Meanwhile, by the definition of $f_{u_{t}}$, the following expressions can be obtained:
\begin{equation}
\label{eq-f}
\begin{aligned}
    f_{u_{t}}\left( \mathbf{X}_1 \right) \mathbf{X}_2 
    =&
    (
    -{}^{D}\Tilde{\mathbf{U}}_{t} \mathbf{X}_1 + \mathbf{X}_1 {}^{B}\Tilde{\mathbf{U}}_{t} 
    ) 
    \mathbf{X}_2,
    \\
    \mathbf{X}_1 f_{u_{t}}\left( \mathbf{X}_2 \right) 
    =&
    \boldsymbol{\mathbf{X}}_1
    (
    -{}^{D}\Tilde{\mathbf{U}}_{t} \mathbf{X}_2 + \mathbf{X}_2 {}^{B}\Tilde{\mathbf{U}}_{t} 
    ),
    \\
    \mathbf{X}_1 f_{u_{t}}\left( \mathbf{I}_{d} \right) \mathbf{X}_2 
    =&
    \mathbf{X}_1
    (
    -{}^{D}\Tilde{\mathbf{U}}_{t} \mathbf{I}_d + \mathbf{I}_d {}^{B}\Tilde{\mathbf{U}}_{t}
    )
    \mathbf{X}_2,
    \\
    =&
    - \boldsymbol{\mathbf{X}}_1 {}^{D}\Tilde{\mathbf{U}}_{t} \boldsymbol{\mathbf{X}}_2
    + 
    \boldsymbol{\mathbf{X}}_1 {}^{B}\Tilde{\mathbf{U}}_{t} \boldsymbol{\mathbf{X}}_2
    .
\end{aligned}
\end{equation}
Note that for the system in \eqref{matrix process model}, the group element $\mathbf{I}_d$ becomes $\mathbf{I}_d = \mathbf{I}_9$ with $\mathbf{I}_m$ an $m \times m$ identity matrix.

Combining the equations in \eqref{eq-f}, we have:
\begin{equation}
\begin{aligned}
    &f_{u_{t}}\left( \boldsymbol{\mathbf{X}}_1 \right) \boldsymbol{\mathbf{X}}_2
    +
    \boldsymbol{\mathbf{X}}_1 f_{u_{t}}\left( \boldsymbol{\mathbf{X}}_2 \right)
    -
    \boldsymbol{\mathbf{X}}_1 f_{u_{t}}\left( \mathbf{I}_{d} \right) \boldsymbol{\mathbf{X}}_2
    \\
    =&
    -{}^{D}\Tilde{\mathbf{U}}_{t}  \boldsymbol{\mathbf{X}}_1 \boldsymbol{\mathbf{X}}_2  + \boldsymbol{\mathbf{X}}_1 \boldsymbol{\mathbf{X}}_2 {}^{B}\Tilde{\mathbf{U}}_{t}
    =
    f_{u_{t}} \left( \boldsymbol{\mathbf{X}}_1  \boldsymbol{\mathbf{X}}_2 \right).
\end{aligned}
\end{equation}
Thus, the group affine condition defined in Theorem 1 of~\cite{barrau2016invariant} is met, confirming the deterministic part of the proposed process model is group affine.
\hfill
$\blacksquare$

\subsection{Process Model Discretization}

Since filters are implemented in a discrete-time fashion in real-world applications, the process model in \eqref{matrix process model} needs to be discretized in order to be used during the propagation step.

Let $t_{k}$ denote the time instant of the $k^{\text{th}}$ measurement update with $k \in \mathbb{N}_{+}$.
With abuse of notation, we use $(\cdot)_k$ to represent the value of a variable $(\cdot)$ at $t_{k}$.
Further, the real scalar $\Delta{t}$ denotes the period between two successive measurement updates; i.e., $\Delta{t} := t_{k+1} - t_{k}$.

Since the process model in \eqref{matrix process model} is a differential Sylvester equation~\cite{behr2019solution},
the closed-form solution of the process model has the following form:
\begin{equation} \label{discretization of dynamics}
    \mathbf{X}_{k+1} = 
    {}^{D}\mathbf{Z}_{k}^{-1}
    \mathbf{X}_{k}
    {}^{B}\mathbf{Z}_{k},
\end{equation}
where the matrix ${}^{i}\mathbf{Z}_{k}$ with $i \in \{B, D \}$ is defined as ${}^{i}\mathbf{Z}_{k} := \exp_m( {}^{i}\Tilde{\mathbf{U}}_{k} \Delta{t} )$.
Using the Taylor expansion of the exponential map $\exp_m( {}^{i}\Tilde{\mathbf{U}}_{k} \Delta{t} )$, we have~\cite{10449450}:
\begin{equation} \label{Z_matrix} 
    \begin{split}
        &{}^{i}\mathbf{Z}_{k} 
        = \sum^{\infty}_{n=0} \tfrac{1}{n!} \left({}^{i}\Tilde{\mathbf{U}}_{k}\Delta{t} \right)^{n} = \sum^{\infty}_{n=0} \tfrac{\Delta t^n}{n!}\begin{bmatrix}
        \left[{}^{i}\Tilde{\boldsymbol{\omega}}_{t} \right]_{\times}& {}^{i}\Tilde{\mathbf{a}}_{t} &\mathbf{0}_{3,1}
            \\
            \mathbf{0}_{1,3} & 0 & 1 
            \\
            \mathbf{0}_{1,3} & 0 & 0
        \end{bmatrix}^{n}
        \\
        &=
        {\small
        \begin{bmatrix}
            \sum\limits^{\infty}_{n=0} \tfrac{1}{n!}\left(({}^{i}\mathbf{z}_{k})^{\wedge} \Delta{t} \right)^{n} 
           &
           \sum\limits^{\infty}_{n=0} \tfrac{1}{(n+1)!}\left(({}^{i}\mathbf{z}_{k})^{\wedge} \Delta{t} \right)^{n} \mathbf{e} \Delta{t}
            \\
            \mathbf{0}_{1,4} &1
        \end{bmatrix}
        },
    \end{split}
\end{equation}
where ${}^{i}\mathbf{z}_{k} := [({}^{i}\boldsymbol{\omega}_{k})^\transpose ~ ({}^{i}\mathbf{a}_{t})^\transpose ]^\transpose $ and $\mathbf{e} := [\mathbf{0}_{1,3} ~1]^\transpose $.
Note that 
\begin{equation}
    \begin{split}
        \begin{bmatrix}
            ({}^{i}\mathbf{z}_{k})^{\wedge} &\mathbf{e} 
            \\
            \mathbf{0}_{1,4} &0
        \end{bmatrix}
        \end{split}:={}^{i}\Tilde{\mathbf{U}}_{k}.
\end{equation}

As shown in~\cite{sola2017quaternion}, the closed-form expression of ${}^{i}\mathbf{Z}_{k}$ is given by:
\begin{equation*}\small \label{Z_k_t}
    {}^{i}\mathbf{Z}_{k} = 
    \begin{bmatrix}
        \boldsymbol{\Gamma}_{0}({}^{i}\boldsymbol{\omega}_{k} \Delta{t}) &\boldsymbol{\Gamma}_{1}({}^{i}\boldsymbol{\omega}_{k} \Delta{t}) {}^{i}\mathbf{a}_{k} &\boldsymbol{\Gamma}_{2}({}^{i}\boldsymbol{\omega}_{k} \Delta{t}) {}^{i}\mathbf{a}_{k}\Delta{t}^{2}
        \\
        \mathbf{0}_{1,3} &1 &\Delta{t}
        \\
        \mathbf{0}_{1,3} &0 &1
    \end{bmatrix},
\end{equation*}
where $\boldsymbol{\Gamma}_{m}(\boldsymbol{\phi}):= \sum^{\infty}_{n=0} \tfrac{1}{(n+m)!} \left[\boldsymbol{\phi}\right]_{\times}^{n}$ with $m \in \{0,1,2 \}$.
According to~\cite{sola2017quaternion}, the closed-form expressions for $\boldsymbol{\Gamma}_{m}(\boldsymbol{\phi})$ can be obtained as:
\begin{equation*} 
\label{gamma}
\begin{split}
        &\boldsymbol{\Gamma}_{0}(\boldsymbol{\phi}) 
        =
        \mathbf{I}_{3} 
        + 
        \tfrac{\sin{\| \boldsymbol{\phi} \|}}{\| \boldsymbol{\phi} \|} \left[\boldsymbol{\phi}\right]_{\times} 
        + 
        \tfrac{1 - \cos{\| \boldsymbol{\phi} \|}}{\| \boldsymbol{\phi} \|^{2}} \left[\boldsymbol{\phi}\right]_{\times}^{2},
        \\
        &
        \begin{split}
        \boldsymbol{\Gamma}_{1}(\boldsymbol{\phi}) 
        &=\mathbf{I}_{3} + \tfrac{1-\cos{\| \boldsymbol{\phi} \|}}{\| \boldsymbol{\phi} \|^{2}} \left[\boldsymbol{\phi}\right]_{\times} + \tfrac{\| \boldsymbol{\phi} \| - \sin{\| \boldsymbol{\phi} \|}}{\| \boldsymbol{\phi} \|^{3}} \left[\boldsymbol{\phi}\right]_{\times}^{2}, \\
        \boldsymbol{\Gamma}_{2}(\boldsymbol{\phi}) &= \tfrac{1}{2}\mathbf{I} + \tfrac{\| \boldsymbol{\phi} \| - \sin{\| \boldsymbol{\phi} \|}}{\| \boldsymbol{\phi} \|^{3}} \left[ \boldsymbol{\phi} \right]_{\times} + \tfrac{\| \boldsymbol{\phi} \|^{2} + 2\cos{\| \boldsymbol{\phi} \|} - 2 }{2\| \boldsymbol{\phi} \|^{4}} \left[ \boldsymbol{\phi} \right]_{\times}^{2}.
    \end{split}
\end{split}
\vspace{-0.05 in}
\end{equation*}

Given the expression of ${}^{i}\mathbf{Z}_{k}$, we can use \eqref{discretization of dynamics} to discretize the process model and propagate the estimated state $\mathbf{\Bar{X}}_t$ during the propagation step of the filter as explained later.

\subsection{Measurement Model}

Following the notational convention in Sec. I-B, we use ${}^{D}\mathbf{p}^{DF}_{t}$ to denote the robot's stance foot position relative to the dynamic ground frame $\{ D \}$, expressed in $\{ D \}$.
For brevity, we define $\mathbf{d}_{t} := {}^{D}\mathbf{p}^{DF}_{t}$.

When the robot's foot has static contact with the ground of the non-inertial environment (i.e., no foot slipping or rolling on the ground), the foot velocity satisfies
\begin{equation}
    \frac{d}{dt} (\mathbf{d}_{t}) = \mathbf{0}_{3,1}.
    \label{zero foot velocity}
\end{equation}
We use this kinematic property to derive the measurement model of the proposed filter.

Using the kinematics relationship associated with the leg odometry, we obtain:
\begin{equation}  
\mathbf{d}_{t} - \mathbf{p}_{t} =  \mathbf{R}_{t}{s}(\mathbf{q}_{t}).
\label{dp}
\end{equation}
Taking the first time derivative of both sides of \eqref{dp} gives:
\begin{equation} \label{v_foot 01} 
    \begin{split}
        &
        \begin{split}
            \frac{d}{dt}\left(\mathbf{d}_{t} - \mathbf{p}_{t} \right)
            = &\left(\mathbf{R}_{t} \left[{}^{B}\boldsymbol{{\omega}}_{t}\right]_{\times} - \left[{}^{D}\boldsymbol{{\omega}}_{t}\right]_{\times} \mathbf{R}_{t} \right) {s}(\mathbf{q}_{t}) 
            \\
            &+ \mathbf{R}_{t} \mathbf{J} (\mathbf{q}_t) \Dot{\mathbf{q}}_{t},
        \end{split}
    \end{split}
\end{equation}
where $\mathbf{J}(\mathbf{q}_t) = \frac{\partial {s}(\mathbf{q}_{t})}{\partial \mathbf{q}_{t}}$ is the Jacobian of leg odometry $s(\mathbf{q}_t)$, and $\Dot{\mathbf{q}}_{t} $ is the time derivative of the joint angle $\mathbf{q}_{t}$.  

Based on \eqref{zero foot velocity} and \eqref{v_foot 01}, we obtain the observation as:
\begin{equation} \label{foot velocity 01}
    \mathbf{y}_{t} = h(\mathbf{X}_{t}) + \mathbf{n}_{f}, 
\end{equation}
where
\begin{equation*}
\begin{gathered}
    \mathbf{y}_t = \left[{}^{B}\Tilde{\boldsymbol{{\omega}}}_{t}\right]_{\times} {s}(\Tilde{\mathbf{q}}_{t}) +  \mathbf{J}\Dot{\Tilde{\mathbf{q}}}_{t},
    \\
    h(\mathbf{X}) = \mathbf{R}_{t}^\transpose \left( \left[{}^{D}\Tilde{\boldsymbol{{\omega}}}_{t}\right]_{\times} \mathbf{R}_{t}{s}(\Tilde{\mathbf{q}}_{t}) - \mathbf{v}_{t} + \left[{}^{D}\Tilde{\boldsymbol{{\omega}}}_{t}\right]_{\times}\mathbf{p}_{t}\right),
\end{gathered}
\end{equation*}
and $\mathbf{n}_{f}$ is the lumped white Gaussian noise of the uncertainty in the encoder reading $\tilde{\mathbf{q}}_t$ and foot slippage on the ground.

The deterministic portion of the measurement model in \eqref{foot velocity 01} does not satisfy the right-invariant observation form, which is defined as $\mathbf{y}_t = \mathbf{X}_t^{-1} \mathbf{b}$ with a known vector $\mathbf{b}$~\cite{barrau2016invariant}.
Thus, the log-error equation associated with the proposed measurement model does not enjoy the attractive properties of an invariant observation and is thus not necessarily independent of state trajectories or exactly linear for the deterministic case.

Instead, we 
linearize the measurement model as follows:
\begin{equation} \label{approx H matrix}
\mathbf{H}_{t}\boldsymbol{\xi}_{t}+\mathrm{h.o.t}\left(\boldsymbol{\xi}_{t} \right) := {h}\left( \Bar{\mathbf{X}}_{t} \right) - {h}\left( \mathbf{X}_{t}\right).
\end{equation}
Because $\boldsymbol{\eta}_{t} \approx \mathbf{I}_{d} + \boldsymbol{\xi}^{\wedge}_{t}$, the relationships between the true and estimated states can be derived as: 
\begin{equation} 
\label{RVP}
    \begin{gathered}
        \Bar{\mathbf{R}}_{t}\mathbf{R}_{t}^\transpose \approx  
        \mathbf{I}_{3} + \left[\boldsymbol{\xi}^{R}_{t} \right]_{\times} 
        ,
        \\
        \Bar{\mathbf{v}}_{t} - \Bar{\mathbf{R}}_{t}\mathbf{R}_{t}^\transpose \mathbf{v}_{t} \approx \boldsymbol{\xi}^{v}_{t}, \quad \Bar{\mathbf{p}}_{t} - \Bar{\mathbf{R}}_{t}\mathbf{R}_{t}^\transpose \mathbf{p}_{t}\approx \boldsymbol{\xi}^{p}_{t}
        ,
    \end{gathered}
\end{equation}
where the vectors $\boldsymbol{\xi}_t^{R}$, $\boldsymbol{\xi}_t^{v}$, and $\boldsymbol{\xi}_t^{p}$ are defined such that:
\begin{equation}
    \boldsymbol{\xi}_t^{\wedge} =:
    \begin{bmatrix}
        \left[\boldsymbol{\xi}_t^{R} \right]_{\times} &\boldsymbol{\xi}_t^{v} &\boldsymbol{\xi}_t^{p}
        \\
        \mathbf{0}_{1,3} &0 &0
        \\
        \mathbf{0}_{1,3} &0 &0
    \end{bmatrix}.
\end{equation}

By differentiating \eqref{approx H matrix}, applying the first-order approximation in \eqref{RVP} to the resulting equation, and then dropping the higher-order terms in the equation, we obtain the expression of the update matrix $\mathbf{H}_{t}$ as:
\begin{equation}  
\label{H1_case2} 
        \mathbf{H}_{t}=
        \begin{bmatrix}
            \mathbf{c}_t  &-\Bar{\mathbf{R}}_{t}^\transpose & \Bar{\mathbf{R}}_{t}^\transpose \left[{}^{D}\Tilde{\boldsymbol{{\omega}}}_{t}\right]_{\times}
        \end{bmatrix}
\end{equation}
with
$
\mathbf{c}_{t} := \Bar{\mathbf{R}}_{t}^\transpose \left[\left[{}^{D}\Tilde{\boldsymbol{{\omega}}}_{t}\right]_{\times} \Bar{\mathbf{R}}_{t} s(\Tilde{\mathbf{q}}) \right]_{\times} -  \Bar{\mathbf{R}}_{t}^\transpose \left[{}^{D}\Tilde{\boldsymbol{{\omega}}}_{t}\right]_{\times} \left[ \Bar{\mathbf{R}}_{t} s(\Tilde{\mathbf{q}}) \right]_{\times} 
$
$+\Bar{\mathbf{R}}_{t}^\transpose \left[\left[{}^{D}\Tilde{\boldsymbol{{\omega}}}_{t}\right]_{\times} \Bar{\mathbf{p}}_{t} \right]_{\times} - \Bar{\mathbf{R}}_{t}^\transpose \left[{}^{D}\Tilde{\boldsymbol{{\omega}}}_{t}\right]_{\times} \left[ \Bar{\mathbf{p}}_{t} \right]_{\times}
$.

\section{Filter Design}

This section introduces the propagation and measurement update steps of the proposed InEKF. 

\subsection{Propagation Step}

\subsubsection{Error Dynamics of Process Model}

By the methodology of InEKF, the right-invariant estimation error $\boldsymbol{\eta}_t$ between the state $\mathbf{X}_t$ and its estimate $\Bar{\mathbf{X}}_t$ is defined as: 
\begin{equation*}
    \boldsymbol{\eta}_t=\Bar{\mathbf{X}}_{t}\mathbf{X}_{t}^{-1}.
\end{equation*}

Thanks to the group-affine property of the proposed process model \cite{barrau2016invariant}, the right-invariant error dynamics in the absence of noise are independent of state trajectories and exactly log-linear in the deterministic case, which is derived next.

Because the process model is group affine, the dynamics of the right-invariant error $\boldsymbol{\eta}_t$ is given by~\cite{barrau2016invariant}: 
\begin{equation}  
\label{deterministic dynamics}
    \begin{split}
        \frac{d}{dt}\boldsymbol{\eta}_{t}
        = g_{u_{t}}(\boldsymbol{\eta}_{t}) + \left(\Bar{\mathbf{X}}_{t} ({}^{B}\mathbf{w}_{t})^{\wedge} \Bar{\mathbf{X}}_{t}^{-1} \right) \boldsymbol{\eta}_{t} + ({}^{D}\mathbf{w}_{t})^{\wedge} \boldsymbol{\eta}_{t},
    \end{split}
\end{equation}
where $g_{u_{t}}(\boldsymbol{\eta}_{t}):= f_{u_{t}}(\boldsymbol{\eta}_{t}) - \boldsymbol{\eta}_{t} f(\mathbf{I}_{d}) $.
Note that by the InEKF theory, the deterministic part of the right-invariant error ($\frac{d}{dt}\boldsymbol{\eta}_{t}= g_{u_{t}}(\boldsymbol{\eta}_{t}) $) are state trajectory independent and accordingly independent of estimation errors.

By using the first-order approximation $\boldsymbol{\eta}_{t}=\exp{(\boldsymbol{\xi}_{t})}\approx \mathbf{I}_{d}+\boldsymbol{\xi}^{\wedge}_{t}$, we linearize \eqref{deterministic dynamics} to yield:
\begin{equation} \label{linearized error dynamics}
   g_{u_t}(\exp({\boldsymbol{\xi}_{t}})) =: (\mathbf{A}_{t}\boldsymbol{\xi}_{t})^{\wedge} + \mathrm{h.o.t}(\lVert \boldsymbol{\xi}_{t} \rVert) \approx (\mathbf{A}_{t}\boldsymbol{\xi}_{t})^{\wedge},
\end{equation}
where $\mathrm{h.o.t}(\cdot)$ represents the higher-order terms of $(\cdot)$.

Then, the linearized log-error dynamics becomes:
\begin{equation} \label{log linearize}
    \begin{split}
        \frac{d}{dt}\boldsymbol{\xi}_{t} =  \mathbf{A}_{t} \boldsymbol{\xi}_{t} + \mathrm{Ad}_{\Bar{\mathbf{X}}_{t}} {}^{B}\mathbf{w}_{t} + {}^{D}\mathbf{w}_{t}.
    \end{split}
\end{equation}

Since the deterministic part of the right-invariant error equation are state trajectory independent, the logarithmic error dynamics are naturally independent of state trajectories in the absence of noise, as indicated by \eqref{log linearize}.
Further, the linear error equation \eqref{log linearize} is exact in the absence of noise.

\vspace{+0.05 in}
\noindent \textbf{Proposition 2:} 
{\it In the absence of the noise terms in the stochastic process model \eqref{matrix process model}, the deterministic portion of the logarithmic error dynamics \eqref{log linearize}, i.e., $\frac{d}{dt}\boldsymbol{\xi}_{t} =  \mathbf{A}_{t} \boldsymbol{\xi}_{t}$, are exact and represent the true error dynamics during propagation.}

\vspace{+0.05 in}
\noindent {\it Proof}:
By Proposition 1, the deterministic part of the process model \eqref{matrix process model} is group affine.
Then, by Theorem 2 in~\cite{barrau2016invariant}, the logarithmic error dynamics in the absence of noise ${}^{B}\Tilde{\boldsymbol{{\omega}}}_{t}$ and ${}^{D}\Tilde{\boldsymbol{{\omega}}}_{t}$ are exact, which completes the proof. 
\hfill $\blacksquare$

By Proposition 2, the linear equation in \eqref{log linearize} is the exact dynamics of the error $\boldsymbol{\xi}_t$ in the absence of noise terms.
Such linearity is rare for nonlinear process models, and holds here because the deterministic portion of the process model is group affine for the deterministic case, as stated in the proof.

The log-error equation in \eqref{log linearize} is used to form the propagation step of the proposed InEKF, and the advantage of its exactness is illustrated via experiment results.

To obtain the matrix $\mathbf{A}_{t}$, we substitute the right-invariant error dynamics \eqref{deterministic dynamics} into \eqref{linearized error dynamics}, which yields:
\begin{equation} 
    \begin{split}
        {g}_{u_{t}}(\exp(\boldsymbol{\xi}_{t})) &\approx f( \mathbf{I}_{d}+\boldsymbol{\xi}^{\wedge}_{t}) - \left( \mathbf{I}_{d}+\boldsymbol{\xi}^{\wedge}_{t} \right) f(\mathbf{I}_{d}) 
        \\
        &=
        \begin{bmatrix}
            -\left[{}^{D}\tilde{\boldsymbol{\omega}}_{t}\right]_{\times}\boldsymbol{\xi}^{R}_{t}
            \\
            -\left[{}^{D}\tilde{\mathbf{a}}_{t}\right]_{\times}\boldsymbol{\xi}^{R}-\left[{}^{D}\tilde{\boldsymbol{\omega}}_{t}\right]_{\times}\boldsymbol{\xi}^{v}_{t}
            \\
            \boldsymbol{\xi}^{v}_{t} -\left[{}^{D}\tilde{\boldsymbol{\omega}}_{t}\right]_{\times}\boldsymbol{\xi}^{p}_{t}
        \end{bmatrix}^{\wedge}.
    \end{split}
\end{equation}
Then, based on \eqref{linearized error dynamics}, we obtain the matrix $\mathbf{A}_{t}$ as:
\begin{equation}
    \mathbf{A}_{t}=
    \begin{bmatrix}
        -\left[{}^{D}\Tilde{\boldsymbol{{\omega}}}_{t}\right]_{\times} & \mathbf{0}_{3,3} & \mathbf{0}_{3,3} 
        \\
       -\left[{}^{D}\Tilde{\mathbf{{a}}}_{t}\right]_{\times} &-\left[{}^{D}\Tilde{\boldsymbol{{\omega}}}_{t}\right]_{\times} &  \mathbf{0}_{3,3} 
        \\
        \mathbf{0}_{3,3} & \mathbf{I}_{3} &-\left[{}^{D}\Tilde{\boldsymbol{{\omega}}}_{t}\right]_{\times} 
    \end{bmatrix}.
\end{equation}


\subsubsection{State and Covariance Propagation}

Between two successive instants of measurement updates, i.e., $t \in [t_k, t_{k+1})$ ($k \in \mathbb{N}_+$), the estimated state $\Bar{\mathbf{X}}_t$ can be propagated~\cite{10449450} using the discretized process model in \eqref{discretization of dynamics}:
\begin{equation*} 
    \Bar{\mathbf{X}}_{k+1} = {}^{D}\mathbf{Z}^{-1}_{k} \Bar{\mathbf{X}}_{k} {}^{B}\mathbf{Z}_{k}.
\end{equation*}

By the theory of the standard Kalman filtering for continuous-time systems, the covariance matrix $\mathbf{P}_{t}$ is propagated based on the following Riccati equation~\cite{maybeck1982stochastic} associated with the linearized log-error equation in \eqref{log linearize}:
\begin{equation}
    \label{eq:riccati}
    \frac{d}{dt}\mathbf{P}_{t} = \mathbf{A}_{t} \mathbf{P}_{t} + \mathbf{P}_{t} \mathbf{A}_{t}^\transpose + \Bar{\mathbf{Q}}_{t},
\end{equation}
where $\Bar{\mathbf{Q}}_{t}$ is the process noise covariance defined as:
\begin{equation}
    \Bar{\mathbf{Q}}_{t} = \mathrm{Ad}_{\Bar{\mathbf{X}}_{t}} \mathrm{Cov}({}^{B}\mathbf{w}_{t}) \mathrm{Ad}_{\Bar{\mathbf{X}}_{t}}^\transpose + \mathrm{Cov}({}^{D}\mathbf{w}_{t})
\end{equation}
with $\mathrm{Cov}({}^{i}\mathbf{w}_{t})$ the covariance of ${}^{i}\Tilde{\mathbf{w}}_{t}$ ($i \in \{B, D\}$).

In filter implementation, the discrete version of the Riccati equation \eqref{eq:riccati} is used for covariance propagation.



\subsection{Update Step}
Based on the measurement model introduced in Sec. III-C, the update equations of the proposed InEKF are:
\begin{equation}
\label{update}
\begin{gathered}
        \Bar{\mathbf{X}}_{t}^{+} = \exp \left( \mathbf{K}_{t} (\mathbf{y}_{t} - h(\Bar{\mathbf{X}}_{t}))\right) \Bar{\mathbf{X}}_{t}~\text{and}
        \\
        \mathbf{P}_{t}^{+} = \left(\mathbf{I}_{9}  \mathbf{K}_{t}\mathbf{H}_{t} \right)\mathbf{P}_{t}  \left(\mathbf{I}_{9} - \mathbf{K}_{t}\mathbf{H}_{t} \right)^\transpose +  \mathbf{K}_{t}\mathbf{N}_{t}\mathbf{K}_{t}^\transpose ,
\end{gathered}
\end{equation}
where $\Bar{\mathbf{X}}_{t}^{+}$ and $\mathbf{P}_{t}^{+}$ are the updated values of the state estimate $\Bar{\mathbf{X}}_{t}$ and covariance matrix $\mathbf{P}_{t}$, respectively, $\mathbf{K}_{t}$ is the Kalman gain,
and $\mathbf{N}_{t}$ is the measurement covariance matrix.
The Kalman gain $\mathbf{K}_{t}$ is given by:
$\mathbf{K}_{t} = \mathbf{P}_{t}\mathbf{H}_{t}^\transpose \mathbf{S}_{t}^{-1}$,
$\mathbf{S}_{t} := \mathbf{H}_{t}\mathbf{P}_{t}^{-}\mathbf{H}_{t}^\transpose + \mathbf{N}_{t}$,
and
$\mathbf{N}_{t} := \Bar{\mathbf{R}}_{t} \mathrm{Cov}(\mathbf{n}_f) \Bar{\mathbf{R}}_{t}^\transpose $.

\section{Observability Analysis}
\label{sec:Observability analysis}

This section reports the observability analysis of the proposed filter system.

Assuming that IMU measurements are constant over the propagation step on $\left[t_{k}, t_{k+1} \right)$, the matrix $\mathbf{A}_{k}$ is constant.
Thus, the discrete-time state-transition matrix, denoted as $\boldsymbol{\Phi}_{k}$, is given by\cite{robocentric_VIO}:
\begin{equation*}
    \begin{split}
       \boldsymbol{\Phi}_{k}
        &= \mathrm{exp}_{m}(\mathbf{A}_{k}\Delta{t})        
        \begin{bmatrix}
            \boldsymbol{\phi}_{11} & \mathbf{0}_{3,3} &\mathbf{0}_{3, 3}
            \\
            \boldsymbol{\phi}_{21} & \boldsymbol{\phi}_{22} &\mathbf{0}_{3, 3}
            \\
            \boldsymbol{\phi}_{31}  &\boldsymbol{\phi}_{32} &\boldsymbol{\phi}_{33}
        \end{bmatrix},
    \end{split}
\end{equation*}
where
\vspace{-0.05 in}
\begin{equation}
    \begin{split}
        &\boldsymbol{\phi}_{11} = \boldsymbol{\phi}_{22} = \boldsymbol{\phi}_{33} = \mathrm{exp}_{m}(-[{}^{D}\boldsymbol{\omega}_{k}]_{\times} \Delta{t}),
        \\
        &\boldsymbol{\phi}_{21} = -[{}^{D}\mathbf{a}_{k}]_{\times} \mathrm{exp}_{m}(-[{}^{D}\boldsymbol{\omega}_{k}]_{\times} \Delta{t})\Delta{t},
        \\
        &\boldsymbol{\phi}_{31} = -\frac{1}{2}[{}^{D}\mathbf{a}_{k}]_{\times} \mathrm{exp}_{m}(-[{}^{D}\boldsymbol{\omega}_{k}]_{\times} \Delta{t}) \Delta{t}^{2},
        \\
        &\boldsymbol{\phi}_{32} = \mathrm{exp}_{m}(-[{}^{D}\boldsymbol{\omega}_{k}]_{\times} \Delta{t}) \Delta{t} .
    \end{split}
\end{equation}

\begin{figure}
    \centering
    \includegraphics[width=0.8\columnwidth]{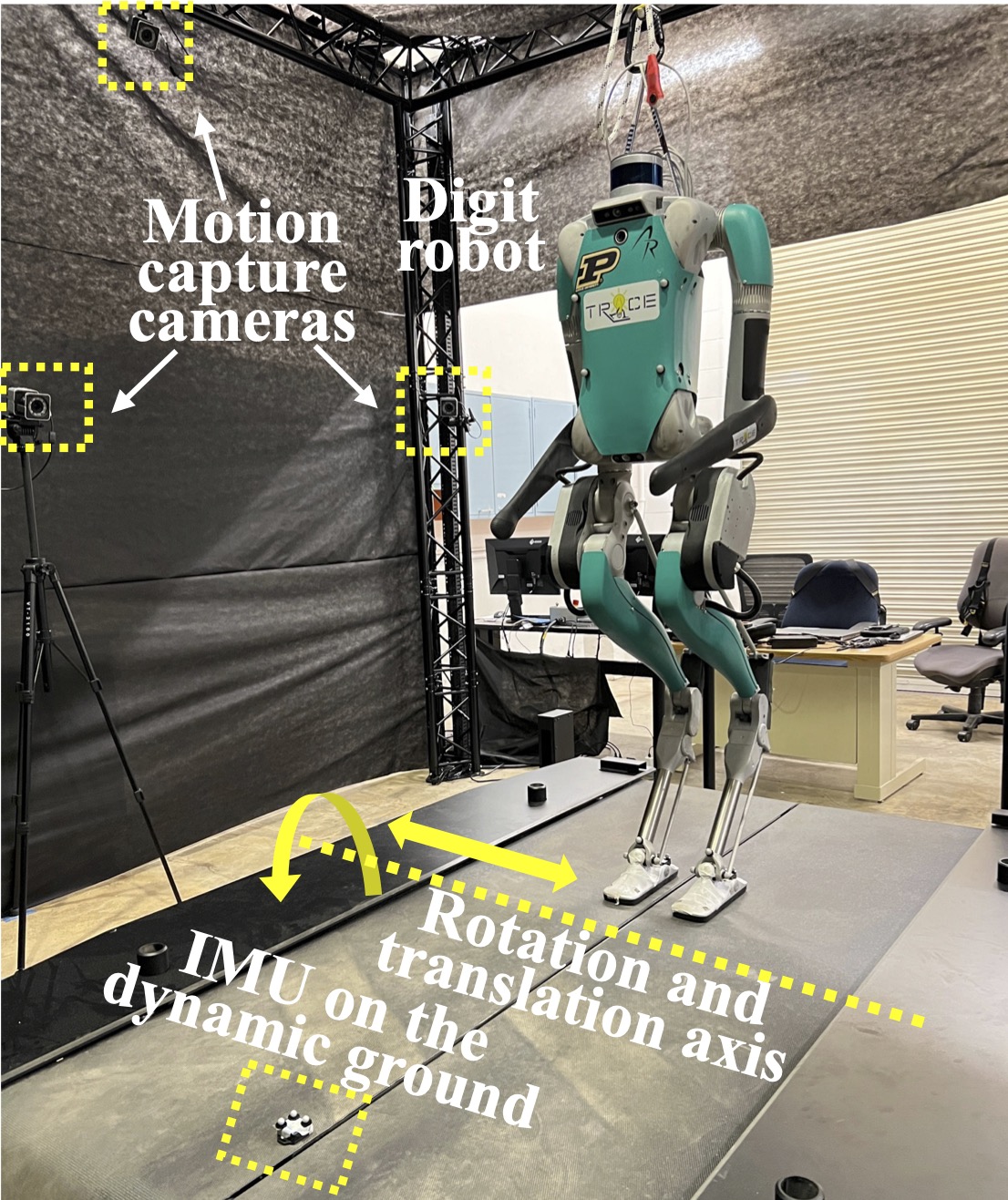}
    \caption{Experimental setup that includes a Digit robot, motion capture cameras, a pitch sway treadmill, and an IMU mounted on the dynamic ground.}
    \label{fig:exp_setup}
\end{figure}

Then, the local observability matrix $\mathcal{O}$ \cite{chen1990local} at the state estimate $\Bar{\mathbf{X}}_k$ is expressed as: 
\begin{equation}
\label{observability}
    \mathcal{O} =
    \begin{bmatrix}
        (\mathbf{H}_{k}^{-})
        \\
        (\mathbf{H}_{k+1}^{-} \boldsymbol{\Phi}_{k}^{+})
        \\
        (\mathbf{H}_{k+2}^{-} \boldsymbol{\Phi}_{k+1}^{+} \boldsymbol{\Phi}_{k}^{+})
        \\
        \vdots
    \end{bmatrix}.
\end{equation}
By definition, $\mathcal{O}$ can be computed as:
\begin{equation} 
\mathcal{O}=
    \begin{bmatrix}
        \mathbf{o}_{11} &-\Bar{\mathbf{R}}_{k}^\transpose & \mathbf{R}^\transpose _{k} \left[{}^{D} \boldsymbol{\omega}_{k} \right]_{\times}
        \\
        \mathbf{o}_{21} &\mathbf{o}_{22} & \mathbf{R}^\transpose _{k+1} \left[ {}^{D}\boldsymbol{\omega}_{k} \right]_{\times} \boldsymbol{\phi}_{33}^{k}
        \\
        \mathbf{o}_{31} &\mathbf{o}_{32} & \Bar{\mathbf{R}}_{k+2}^\transpose [{}^{D}\boldsymbol{\omega}_{k+2}]_{\times}\boldsymbol{\phi}_{33}^{k+1} \boldsymbol{\phi}_{33}^{k}, 
        \\
        \vdots &\vdots &\vdots
    \end{bmatrix},
\end{equation}
where
\begin{equation*} 
    \begin{split}
        &
\begin{aligned}
        \mathbf{o}_{11} &= \mathbf{c}_{k},
\end{aligned}
        \\
        &
        \begin{split}
        \mathbf{o}_{21} 
            &= \mathbf{c}_{k+1} \boldsymbol{\phi}_{11}^{k} -   \Bar{\mathbf{R}}_{k+1}^\transpose \boldsymbol{\phi}_{21}^{k} + \Bar{\mathbf{R}}_{k+1}^\transpose [{}^{D}\boldsymbol{\omega}_{k} ]_{\times} \boldsymbol{\phi}_{31}^{k},
        \end{split}
        \\
       &
        \begin{split}
            \mathbf{o}_{22} &= -\Bar{\mathbf{R}}_{k+1}^\transpose \boldsymbol{\phi}_{22}^{k} + \Bar{\mathbf{R}}_{k+1}^\transpose [{}^{D}\boldsymbol{\omega}_{k} ]_{\times} \boldsymbol{\phi}_{33}^{k},
    \end{split}
        \\
        &
        \begin{split}
        \mathbf{o}_{31} 
            &= \mathbf{c}_{k+2} \boldsymbol{\phi}_{11}^{k+1} \boldsymbol{\phi}_{11}^{k} -\Bar{\mathbf{R}}_{k+2}^\transpose \boldsymbol{\phi}_{21}^{k+1}\boldsymbol{\phi}_{11}^{k} 
            \\
            & \quad - \Bar{\mathbf{R}}_{k+2}^\transpose \boldsymbol{\phi}_{22}^{k+1} \boldsymbol{\phi}_{21}^{k} + \Bar{\mathbf{R}}_{k+2}^\transpose [{}^{D}\boldsymbol{\omega}_{k+2}]_{\times} \boldsymbol{\phi}_{31}^{k+1} \boldsymbol{\phi}_{11}^{k} 
            \\
            & \quad+ \Bar{\mathbf{R}}_{k+2}^\transpose [{}^{D}\boldsymbol{\omega}_{k+2}]_{\times}\boldsymbol{\phi}_{32}^{k+1} \boldsymbol{\phi}_{21}^{k}
            \\
            &\quad + \Bar{\mathbf{R}}_{k+2}^\transpose [{}^{D}\boldsymbol{\omega}_{k+2}]_{\times}\boldsymbol{\phi}_{33}^{k+1} \boldsymbol{\phi}_{31}^{k},
        \end{split}
        \\
       &
        \begin{split}
            \mathbf{o}_{32} &=   -\Bar{\mathbf{R}}_{k+2}^\transpose \boldsymbol{\phi}_{22}^{k+1} \boldsymbol{\phi}_{22}^{k+2} + 2\Bar{\mathbf{R}}_{k+1}^\transpose [{}^{D}\boldsymbol{\omega}_{k} ]_{\times} \boldsymbol{\phi}_{32}^{k+1} \boldsymbol{\phi}_{22}^{k+1}
            \\
            &\quad + \Bar{\mathbf{R}}_{k+1}^\transpose [{}^{D}\boldsymbol{\omega}_{k} ]_{\times} \boldsymbol{\phi}_{33}^{k+1} \boldsymbol{\phi}_{32}^{k+1}.
        \end{split}
    \end{split}
\end{equation*}

To evaluate the observability of each variable of interest, we examine whether the associated column vectors in the observability matrix $\mathcal{O}$ are linearly independent.

From the expression of $\mathcal{O}$, the observability of the state variables depends on the estimated relative orientation $\Bar{\mathbf{R}}_{t}$ as well as the linear acceleration data ${}^{D}\mathbf{a}_{t}$ and angular velocity data ${}^{D}\boldsymbol{\omega}_{t}$ of the dynamic ground.
The estimate $\Bar{\mathbf{R}}_{t}$ is always a non-zero matrix.
Thus, when the ground is rotating and translating (i.e., ${}^{D}\mathbf{a}_{t}\neq \mathbf{0}$ and $ {}^{D}\boldsymbol{\omega}_{t}\neq \mathbf{0}$), all columns of $\mathcal{O}$ are linearly independent.
Accordingly, the robot's relative orientation ${\mathbf{R}}_{t}$, velocity ${\mathbf{v}}_{t}$, and position ${\mathbf{p}}_{t}$ are observable when the ground is moving.

When the ground is stationary, the angular velocity data ${}^{D}\Tilde{\boldsymbol{\omega}}_{t}$ is zero in the absence of sensor noise, and thus the entire third column block becomes zeros, indicating the relative position ${\mathbf{p}}_{t}$ is no longer observable.
However, even when the ground is not moving, the linear acceleration data ${}^{D}\Tilde{\mathbf{a}}_{t}$ remains nonzero because ${}^{D}\Tilde{\mathbf{a}}_{t}$ includes the gravitational acceleration in the vertical direction of the world frame. 
Thus, the third column of $\left[ {}^{D}\mathbf{a}_{t}\right]_{\times}$ is zero, indicating the yaw angle is non-observable when the ground is stationary. 

\begin{figure}[t]
    \centering
    \includegraphics[scale=0.39]{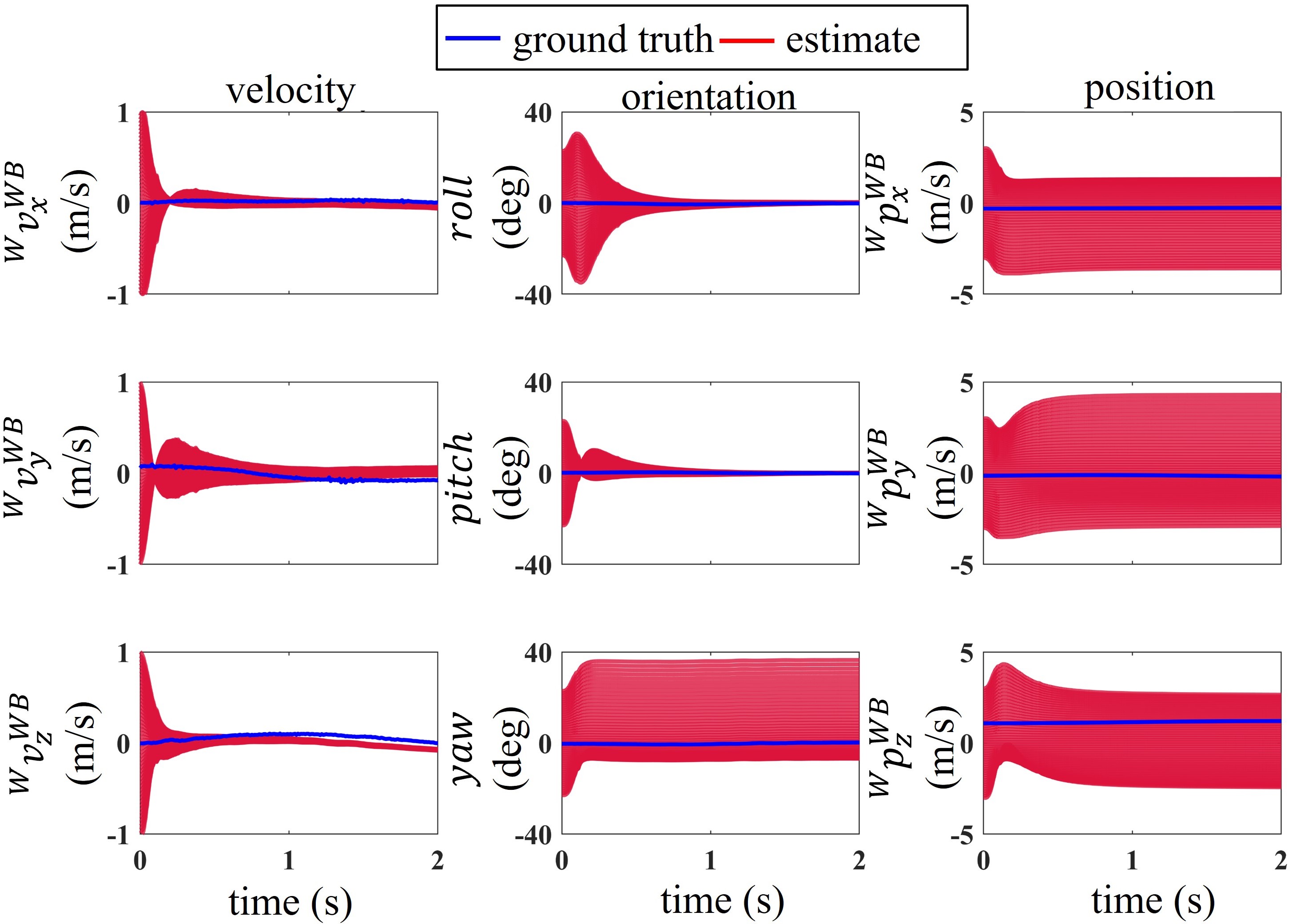}
    \caption{Estimation results of the SRS filter during the transient period.}
    \label{fig:}
\end{figure}

\begin{figure}[t]
    \centering
    \includegraphics[scale=0.39]{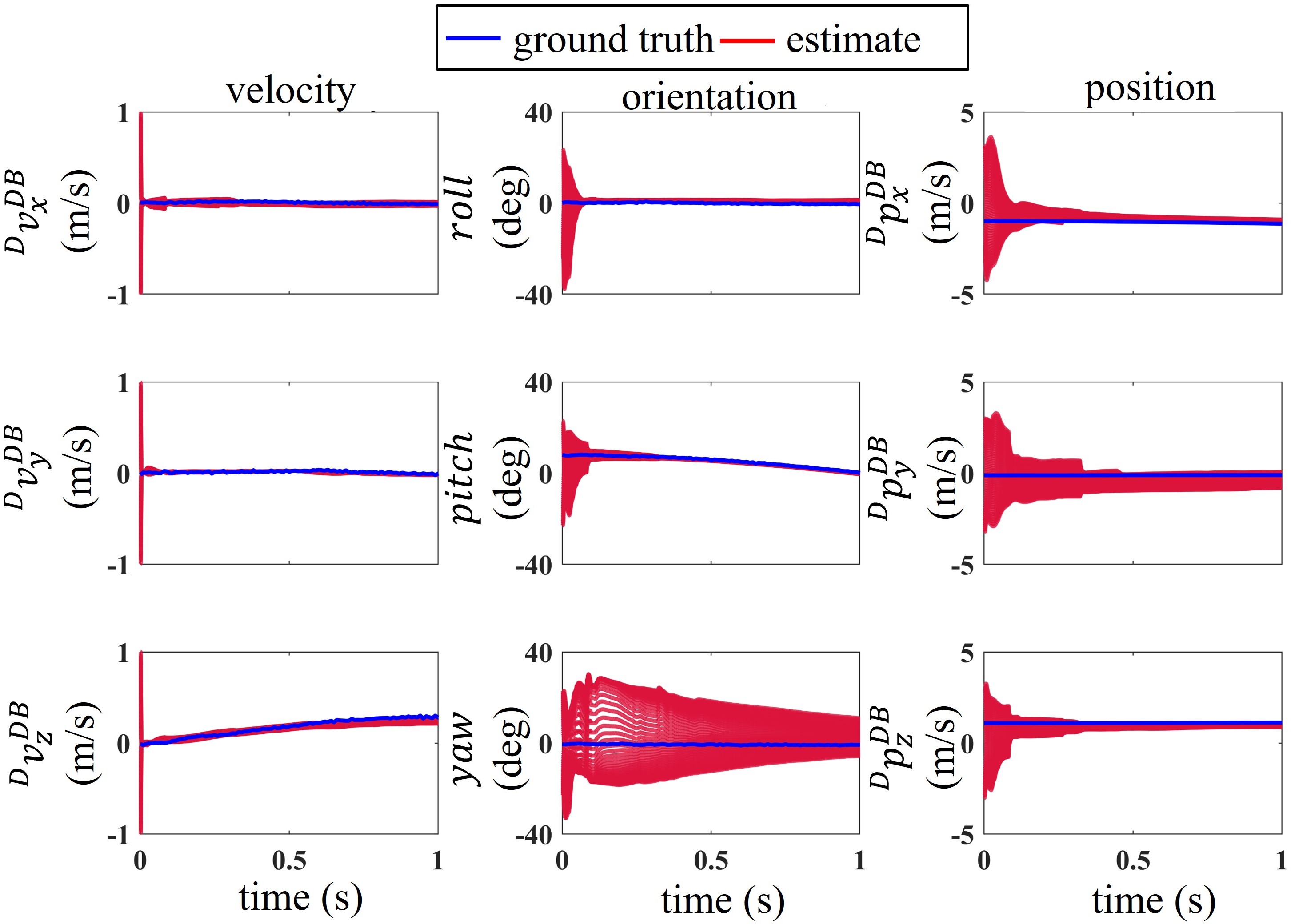}
    \caption{Estimation results of the proposed filter during the transient period.}
    \label{fig:}
\end{figure}

\section{Experimental Results}

The corresponding error equation at the measurement update is omitted for brevity and can be readily obtained based on the definitions of the errors and the update equation in \eqref{update}.

\subsection{Experimental setup}

Hardware experiments on a Digit humanoid robot (Agility Robotics, Inc.) and a Motek M-Gait treadmill are performed to assess the proposed filter (see Fig.~\ref{fig:exp_setup}).
Digit is 1.6 m tall with 6 encoders on each leg.
During experiments, the robot stands on the treadmill.
To simulate the dynamic ground of a non-inertial environment, the treadmill is programmed to simultaneously perform a sinusoidal pitch motion, $10^{\circ}\sin{\frac{\pi t}{2}}$, and a sway motion $0.05\mathrm{m}\cos{\frac{\pi t}{2}}$.

An IMU is attached to the treadmill and returns the angular velocity and linear acceleration of the dynamic ground frame at 200 Hz via Bluetooth.
The robot IMU provides linear acceleration and angular velocity in the robot frame at 500 Hz, and the encoder supplies readings of joint positions at the same rate of 500 Hz.
Additionally, a Vicon motion capture system captures the ground-truth value of the state $\mathbf{X}_t$.


\subsection{Filter Setup}
The proposed filter is compared with an InEKF \cite{hartley2020contact} designed for locomotion on a static, rigid surface (denoted as ``SRS"),
so as to highlight the advantage of explicitly treating the environment/ground motion in the filter formulation.

The key difference between the proposed and the SRS filters is that the SRS filter assumes the ground is stationary.
Accordingly, the SRS filter aims to estimate the robot's absolute base position, orientation, and velocity expressed in the world frame, which is different from the proposed filter.
Although the process models of the two filters are different due to different choices of state variables, both models meet the group-affine property for the deterministic case.
This indicates that both filters obey the attractive property of invariant filtering, such as the exact linearity and state independence of log-error dynamics for the deterministic part of the process model.
Also, the measurement models of both filters exploit the leg odometry, with the SRS filter using a position-based one while the proposed filter relies on a velocity-based one.
However, the baseline filter has a right-invariant measurement model, while the proposed one does not.

\begin{table}[t]
\caption{\small NOISE STANDARD DEVIATION}
\centering
\footnotesize
\begin{tabular}{@{}lcccc@{}}
\toprule
\multirow{2}{*}{Measurement types} & \multicolumn{2}{c}{Robot standing}  \\ 
\cmidrule(r){2-3} \cmidrule(r){4-5}
          & SRS & Proposed  \\
\midrule
Robot linear acc. ($\mathrm{m/s}^{2}$)     &0.3  & 0.1 \\
Ground linear acc. ($\mathrm{rad/s}$)    &NA & 0.1 \\
Robot angular vel. ($\mathrm{m/s}^{2}$)     &0.01 &0.01  \\
Ground angular vel. ($\mathrm{rad/s}$)     &NA &0.01  \\
Encoder reading    & $1^{\circ}$ &0.1 m/s  \\
Contact vel. ($\mathrm{m/s}$)  &0.01  &NA \\
\bottomrule
\end{tabular}
\label{tab:covariance}
\end{table}

The setting of the standard deviation (SD) of both filters is shown in Table \ref{tab:covariance}. The SD values of linear accelerations and angular velocities are obtained from the IMU specifications provided by the manufacturers. All the SD values are individually tuned for the two filters to achieve their respective best performance. Both filters are assessed using the same hardware sensor data under the same range of initial estimation errors. To highlight the proposed InEKF can handle large estimation errors, the initial position, velocity, and orientation errors in each of the $x$-, $y$-, and $z$-directions are uniformly sampled from $[-3, 3]$ m, $[-1, 1]$ m/s, and $ [-23, 23]$ deg. 


\begin{figure}[t]
    \centering
    \includegraphics[scale=0.39]{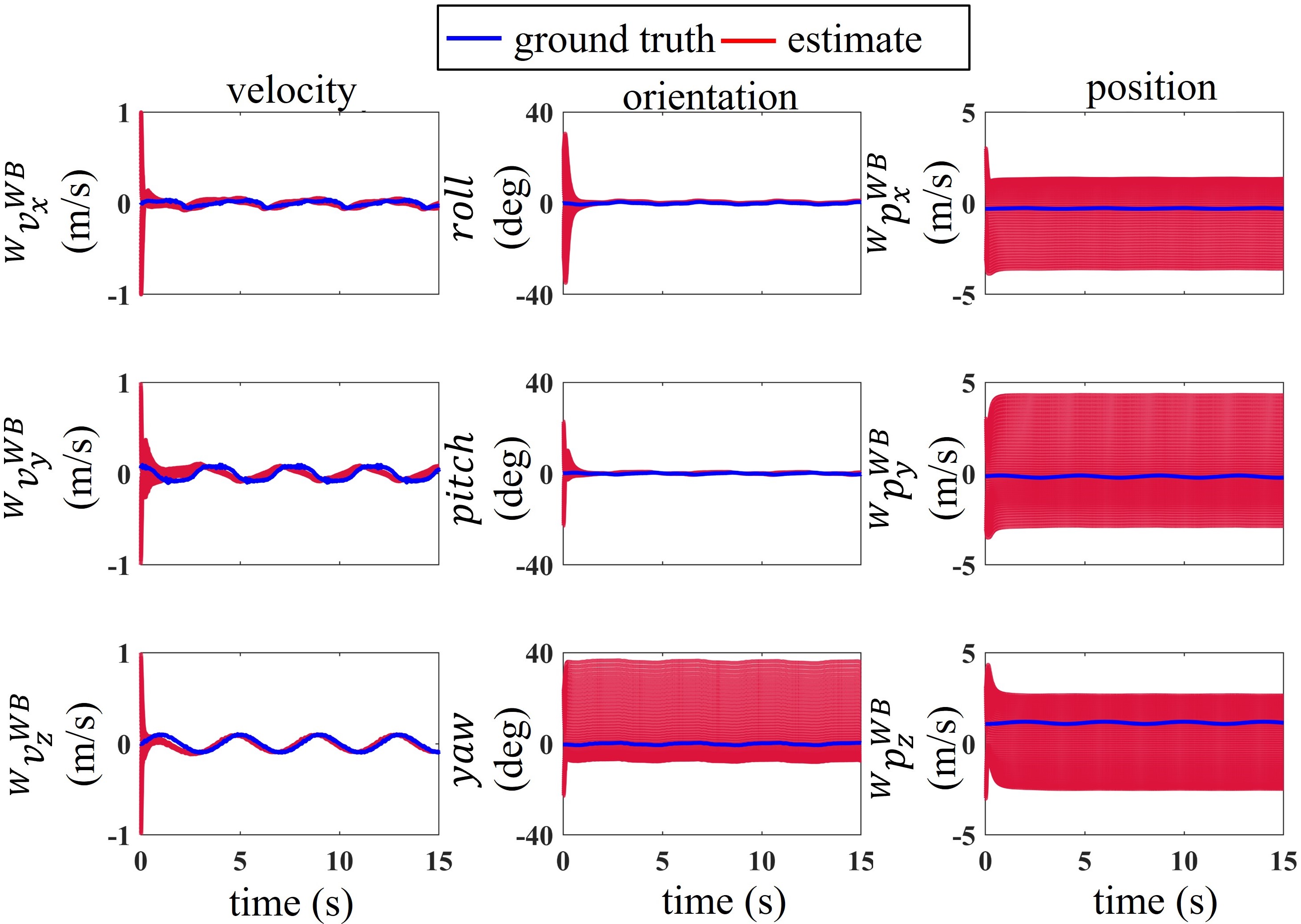}
    \caption{Estimation results of the SRS filter near the steady state.}
    \label{fig:}
\end{figure}

\begin{figure}[t]
    \centering
    \includegraphics[scale=0.39]{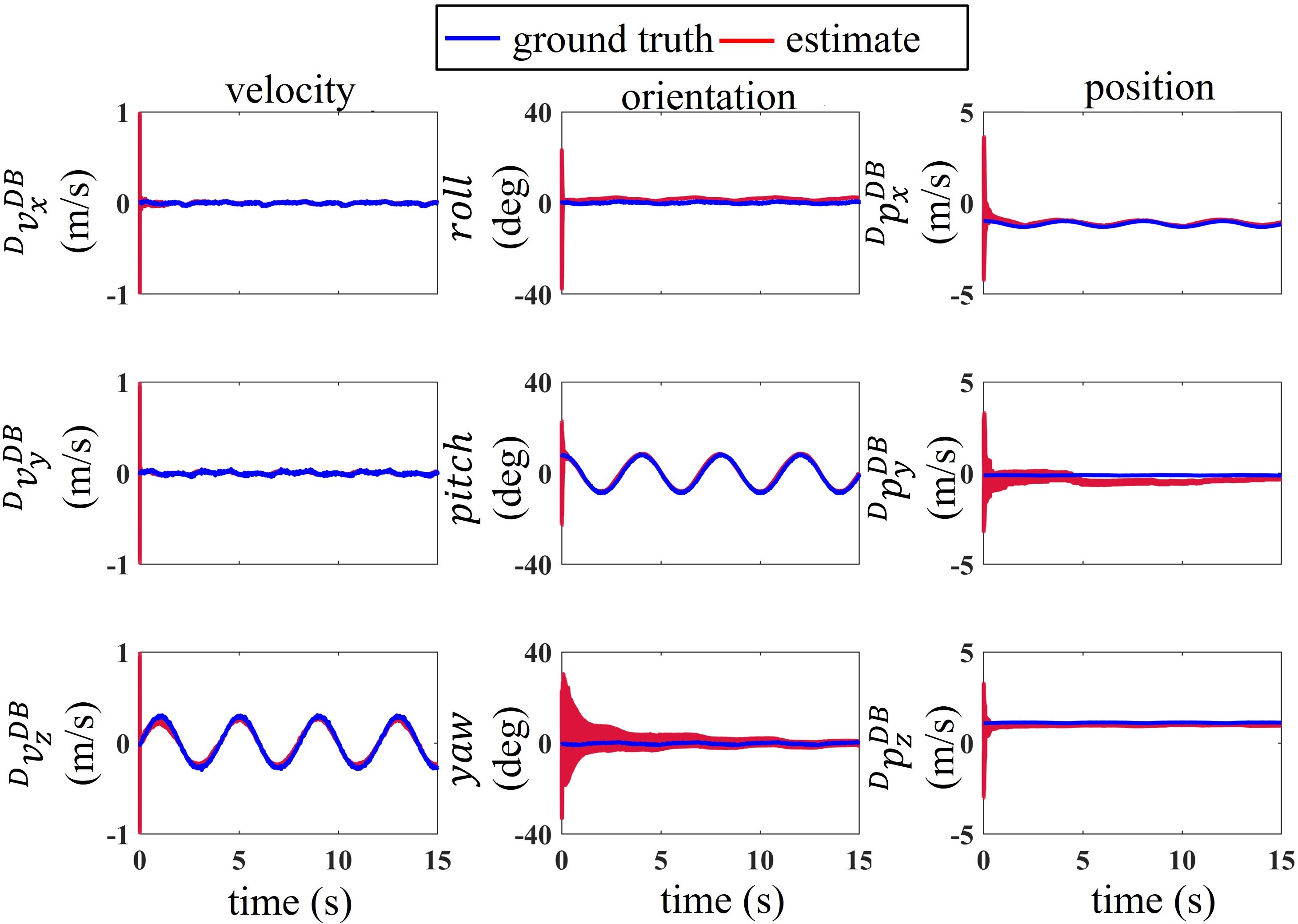}
    \caption{Estimation results of the proposed filter near the steady state.}
    \label{fig:}
\end{figure}

\subsection{Results}

\subsubsection{Convergence Rate}
To illustrate the filter convergence rate, Figs. 3 and 4 display the transient phases of the SRS and the proposed filter on $t \in [0,~1] s$.
The variables displayed in Fig. 3 are the absolute velocity ${}^{W}\mathbf{v}^{WB}$, orientation ${}^{W}\mathbf{R}^{B}$, and position ${}^{W}\mathbf{p}^{WB}$ with respect to the world frame $\{ W \}$ in the $x$-, $y$-, and $z$-directions. 
Meanwhile, the state variables illustrated in Fig. 4 are the relative velocity $\mathbf{v}_t$, orientation $\mathbf{R}_t$, and position $\mathbf{p}_t$ with respect to the dynamic ground frame $\{ D \}$ in $x$-, $y$-, and $z$-directions. 
The plots demonstrate that both filters drive the errors of the base roll, pitch, and velocities close to zero, confirming the observability analysis results from Sec.~\ref{sec:Observability analysis} and previous work \cite{hartley2020contact}.
Both filters show fast error convergence for their respective observable state even under large initial errors, thanks to the advantages of InEKF approaches, including the provable error convergence under the deterministic case.
The proposed filter exhibits a much faster convergence rate than that of the SRS filter due to the explicit treatment of the ground motion.

\subsubsection{Yaw and Position Observability}
Notably, under the proposed filter, the robot's relative base yaw and position also converge to the ground truth, supporting the observability analysis results that they are observable during ground motion.
In contrast, the absolute yaw and position under the SRS filter are not observable as predicted by the previous study~\cite{hartley2020contact}.

\subsubsection{Estimation Accuracy}
Figures 5 and 6 show the steady-state periods that illustrate the filters' final errors.
Table \ref{tab:rmse_values} reports the comparison of the root-mean-square errors (RMSE) between the state estimate and the ground truth for the base position, velocity, and orientation corresponding to ${}^{W}\mathbf{v}^{WB}$, ${}^{W}\mathbf{R}^{B}$, ${}^{W}\mathbf{p}^{WB}$ for the SRS filter, as well as to ${}^{D}\mathbf{v}^{DB}$, ${}^{D}\mathbf{R}^{B}$, ${}^{D}\mathbf{p}^{DB}$ for the proposed filter. 

As the state variables estimated by the two filters have different physical meanings, directly comparing their specific accuracy may not be meaningful.
However, the smaller estimation errors of the proposed method do highlight the need to explicitly consider the ground motion in the state estimation, especially under significant ground motions such as the tested treadmill movement.
Without explicit treatment, the ground motion acts as persistent, significant uncertainties that could notably degrade estimator performance.




\begin{table}[t]
\caption{RMSE COMPARISON}
\centering
\footnotesize 
\begin{tabular}{@{}lcccc@{}}
\toprule
\multirow{2}{*}{State variables} & \multicolumn{2}{c}{Robot standing} \\ 
\cmidrule(r){2-3} \cmidrule(r){4-5}
          & SRS & Proposed\\
\midrule
$v_{x}$ (m/s)      & 0.048 & 0.017 \\
$v_{y}$ (m/s)      & 0.080 & 0.018 \\
$v_{z}$ (m/s)      & 0.041 & 0.040\\
roll (deg)     & 1.889 & 1.886 \\
pitch (deg)    & 1.520 & 0.980  \\
yaw (deg)      & 10.91 & 2.871  \\
$p_{x}$ (m)      &1.145 & 0.283 \\
$p_{y}$ (m)      &1.688 & 0.336 \\
$p_{z}$ (m)      &1.225 & 0.165 \\
\bottomrule
\end{tabular}
\label{tab:rmse_values}
\end{table}

\section{Conclusion}
This paper developed a real-time state estimation approach for legged locomotion inside a non-inertial environment with an unknown motion. 
The process and measurement models underlying the estimator were formulated to explicitly consider the movement of the non-inertial environment.
A minimal suite of proprioceptive sensors and an inertial measurement unit attached to the environment were used to inform the proposed InEKF.
The observability analysis revealed that all state variables (i.e., relative pose and linear velocity) are observable during environment translation and rotation.
Hardware experiment results and comparison with a baseline InEKF demonstrated the fast convergence rate and high accuracy of the proposed filter under various ground motions and substantial estimation errors. 
The proposed system modeling can be readily used in filtering and optimization frameworks beyond InEKF, 
and can be combined with data returned by exteroceptive sensors such as cameras and LiDARs.
Future work includes the study of fully onboard sensing and learning-aided methods to remove the need for an external IMU attached to the moving environment.

{\balance 
\bibliographystyle{IEEEtran}
\bibliography{HeCDC}
}

\end{document}